\documentclass{article}
\usepackage{nips12submit_e,times}

\usepackage{multicol}
\usepackage[bookmarks=true]{hyperref}

\usepackage{graphicx} 
\usepackage{algorithm}
\usepackage{algorithmic}
\usepackage{epsf}
\usepackage{amsmath}
\usepackage{amssymb}
\usepackage{graphicx}
\usepackage{tabularx}
\usepackage{rotating}
\usepackage{url}
\usepackage{mathrsfs}
\def\tps{^{\mathsf{T}}}

\def\Re{{\mathbb R}}

\usepackage{Definitions}

\usepackage{mlapa2}

\title{A Spectral Learning Approach to Range-Only SLAM}

\author{
Byron Boots \\
Machine Learning Department\\
Carnegie Mellon University \\
Pittsburgh, PA 15213\\
\texttt{beb@cs.cmu.edu}
\And
Geoffrey J. Gordon \\
Machine Learning Department\\
Carnegie Mellon University \\
Pittsburgh, PA 15213\\
\texttt{ggordon@cs.cmu.edu}
}

\nipsfinalcopy

\begin{document}

\maketitle

\begin{abstract}
We present a novel spectral learning algorithm for simultaneous
localization and mapping (SLAM) from range data with known correspondences.  This algorithm is an instance of a general spectral system identification framework, from which it inherits several desirable properties, including statistical consistency and no local optima.
Compared with popular batch optimization or multiple-hypothesis tracking
(MHT) methods for range-only SLAM, our spectral
approach offers guaranteed
low computational requirements and good tracking performance.
Compared with popular extended Kalman filter (EKF)
or extended 
information filter (EIF) approaches, and many MHT ones, our 
approach does not need to linearize a transition or measurement model;
such linearizations can cause severe errors in EKFs and
EIFs, and to a lesser extent MHT,
particularly for the highly non-Gaussian posteriors encountered in
range-only SLAM\@.  
We provide a theoretical analysis of our method, including
finite-sample error bounds.  Finally, we demonstrate on a real-world
robotic SLAM problem that our algorithm is
not only theoretically justified, but works well in practice: in a
comparison of multiple methods, the lowest errors come
from a combination of our algorithm with batch
optimization, but our method alone produces nearly as good a result at
far lower computational cost.
\end{abstract}

\section{Introduction}
In range-only SLAM, we are given a sequence of range measurements from
a robot to fixed landmarks, and possibly a matching sequence of
odometry measurements.  We then attempt to simultaneously estimate the
robot's trajectory and the locations of the landmarks.
Popular approaches to range-only SLAM include EKFs and
EIFs~\cite{Kantor2002,Kurth2003,Djugash2008,Djugash2010,Thrun2005},
multiple-hypothesis trackers (including particle filters and multiple
EKFs/EIFs)~\cite{Djugash2005,Thrun2005}, and batch optimization of
a likelihood function~\cite{Kehagias2006}.

In all the above approaches, the most popular representation for a
hypothesis is a list of landmark locations $(m_{n,x},m_{n,y})$ and a
list of robot poses $(x_t,y_t,\theta_t)$.  Unfortunately, both the
motion and measurement models are highly nonlinear in this
representation, leading to computational problems: inaccurate
linearizations in EKF/EIF/MHT and local optima in batch optimization
approaches (see Section~\ref{sec:background} for details).  
Much
work has attempted to remedy this problem, e.g., by changing the
hypothesis representation~\cite{Djugash2010} or by keeping multiple
hypotheses~\cite{Djugash2005,Djugash2010,Thrun2005}.  
While considerable progress has been made, 
none of these methods are ideal;
common difficulties include the need for an extensive initialization
phase, inability to recover
from poor initialization, lack of performance guarantees, or excessive
computational requirements.

We take a very different approach: we formulate
range-only SLAM as a matrix factorization problem, where features of
observations are linearly related to a 4- or 7-dimensional state space. This
approach has several desirable properties. First, we need weaker
assumptions about the measurement model and motion model
than previous approaches to SLAM\@. 
Second, our state space yields a
\emph{linear} measurement model, so we hope to lose less
information during tracking to approximation errors and local optima. Third, our
formulation leads to a simple spectral learning algorithm, based on
a fast and robust singular value decomposition (SVD)---in fact,
our algorithm is an instance of a general spectral system identification framework, from which it inherits desirable guarantees including statistical consistency and no local optima. 
 Fourth, we don't need to worry as much as previous methods about
 errors
such as a consistent bias in odometry,
 or a receiver mounted at a different height from the transmitters: in
 general, we can learn to correct such errors automatically by expanding the dimensionality of our state space.

As we will discuss in Section~\ref{sec:background}, our approach to
SLAM has much in common with spectral algorithms for subspace
identification~\cite{vanoverschee96book,Boots2010b}; unlike these
methods, our
focus on SLAM makes it easy to \emph{interpret} our state space. 
Our approach is also related to factorization-based structure from
motion~\cite{Tomasi92,triggs1996,kanade1998}, as well as to recent
dimensionality-reduction-based methods  for 
localization and mapping~\cite{Shang2003,Biggs05,Ferris2007,Yairi2007}. 

We begin in Section~\ref{sec:background} by reviewing background related to our approach. In Section~\ref{sec:spectralSLAM} we present the basic spectral learning algorithm for range-only SLAM, and discuss how it relates to state space discovery for a dynamical system.  
We conclude in Section~\ref{sec:results} by comparing spectral SLAM to other popular methods for range-only SLAM on real world range data collected from an autonomous lawnmower with time-of-flight ranging radios.

\section{Background}
\label{sec:background}
There are four main pieces of relevant background: first, the
well-known solutions to range-only SLAM using variations of the
extended Kalman filter and batch optimization; second,
recently-discovered spectral approaches to identifying parameters of
nonlinear dynamical systems; third, matrix factorization for finding
structure from motion in video; and fourth, dimensionality-reduction
methods for localization and mapping. Below, we will discuss
the connections among these areas, and show how they can be unified
within a spectral learning framework.

\subsection{Likelihood-based Range-only SLAM}
The standard probabilistic model for range-only
localization~\cite{Kantor2002,Kurth2003} represents 
robot state by a vector $s_t = [x_t, y_t,
\theta_t]\tps$; the robot's (nonlinear) motion and observation models are
\begin{align}
\label{eq:kalman-motion-measurement}
\begin{array}{c}
s_{t+1} = \left [ \begin{array}{c} x_t + v_t \cos(\theta_t)\\
y_t + v_t \sin(\theta_t)\\
\theta_t + \omega_t
\end{array}   \right] + \epsilon_t 
\end{array}
\quad
\begin{array}{c}
d_{t,n} = \sqrt{(m_{n,x} - x_t)^2 + (m_{n,y} - y_t)^2 } + \eta_t
\end{array}
\end{align}
Here $v_t$ is the distance traveled, $\omega_t$ is the orientation
change, $d_{t,n}$ is the estimate of the range from the $n$th landmark
location $(m_{n,x}, m_{n,y})$ to the current location of the robot
$(x_t,y_t)$, and $\epsilon_t$ and $\eta_t$ are noise.  (Throughout
this paper we assume known correspondences, since range sensing
systems such as radio beacons typically associate unique identifiers
with each reading.)

To handle SLAM rather than just localization, we can extend the state
to include landmark positions:
\begin{align}
s_t = [x_t, y_t, \theta_t, m_{1,x}, m_{1,y},  \hdots, m_{N,x}, m_{N,y}]\tps
\end{align}
where $N$ is the number of landmarks. The motion and measurement
models remain the same.  Given this model, we can use any standard
optimization algorithm (such as Gauss-Newton) to fit the unknown robot
and landmark parameters by maximum likelihood.  Or, we can track these
parameters online using EKFs, EIFs, or MHT methods like particle filters.

EKFs and EIFs are a popular solution for localization and mapping
problems: for each new odometry input $a_t = [ v_t, \omega_t]\tps$ and
each new measurement $d_t$, we propagate the estimate of the robot
state and error covariance by linearizing the non-linear motion and
measurement models.
Unfortunately, though,
range-only SLAM is notoriously difficult for EKFs/EIFs: since
range-only sensors are not 
informative enough to completely localize a robot or a landmark from a
small number of readings, nonlinearities are much worse in
range-only SLAM than they are in other applications such
as range-and-bearing SLAM\@.  In particular,
  if we don't have a sharp
prior distribution for landmark positions, then after a few steps,
the exact posterior becomes highly non-Gaussian and multimodal; so,
any Gaussian approximation to the posterior is necessarily inaccurate.
Furthermore, an EKF will generally not even produce the best possible
Gaussian approximation: a good linearization would tell us a lot about
the modes of the posterior, which would be equivalent to solving the
original SLAM problem.  So, practical applications of the EKF to
range-only SLAM attempt to \emph{delay} linearization until enough
information is available, e.g., via an extended initialization phase
for each landmark.  Such delays simply push the problem of finding a
good hypothesis onto the initialization algorithm.

Djugash et al.\ proposed a polar parameterization to more accurately
represent the annular and multimodal distributions typically
encountered in range-only SLAM. The resulting approach is called the
ROP-EKF, and is shown to outperform the ordinary (Cartesian) EKF in
several real-world problems, especially in combination with
multiple-hypothesis tracking~\cite{Djugash2008,Djugash2010}.  However,
the multi-hypothesis ROP-EKF can be much more expensive than an EKF,
and is still a heuristic approximation to the true posterior.

 Instead of the posterior covariance of the state (as used by the EKF),
 the extended information filter (EIF) maintains an estimate of the
 \emph{inverse} covariance.  The two representations are statistically
 equivalent (and therefore have the same failure modes).  But, the
 inverse covariance is often approximately sparse, leading to much more
 efficient approximate computation~\cite{Thrun2005}.

\subsection{Spectral  State Space Discovery and System Identification}
System identification algorithms attempt to \emph{learn} dynamical
system parameters  such as a state space, a dynamics model (motion
model), and an observation model (measurement model) directly from
samples of observations and actions. In the last few years,
\emph{spectral} system identification algorithms have become popular;
these algorithms learn a state space via a spectral decomposition of a
carefully designed matrix of observable features, then find transition
and observation models by linear regressions involving the learned
states. Originally, subspace
identification algorithms were almost exclusively used for linear
system identification~\cite{vanoverschee96book}, but recently, similar
spectral algorithms have been used to learn models of partially
observable nonlinear dynamical systems such as
HMMs~\cite{zhang09,Siddiqi10a} and
PSRs~\cite{rosencrantz04,Boots2010b,Boots2011a,Boots-online-psr}.  
\begin{figure}[!tb]
\begin{center}
\includegraphics[width=.8\columnwidth]{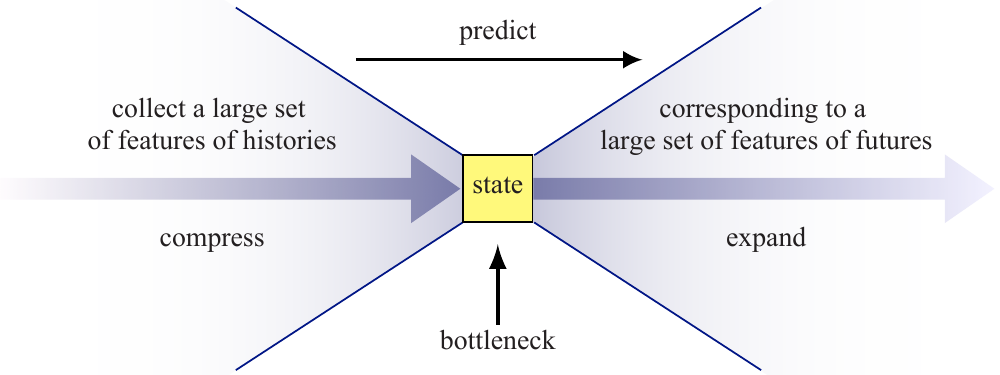}
\end{center}
\caption{A general principle for state space discovery. We can think of state as a \emph{statistic} of history that is minimally  \emph{sufficient} to predict future observations. If the bottleneck is a rank constraint, then we get a \emph{spectral} method.
} \label{fig:bottleneck}
\end{figure}
 All of these spectral algorithms share a strategy for state space
 discovery: they learn a state space via a spectral decomposition of a
 matrix of observations (Figure~\ref{fig:bottleneck}), 
 resulting in a
 linear observation function, and then they learn a model of the
 dynamics in the learned low-dimensional state space. 
This is a
powerful and appealing approach: the resulting 
algorithms are \emph{statistically consistent}, and they are easy to
implement with efficient linear algebra operations.  In contrast, batch
optimization of likelihood (e.g., via 
the popular
expectation maximization (EM) algorithm) is only known to be
consistent if we find the \emph{global} optimum of the likelihood
function---typically an impractical requirement.

As we will see in Section~\ref{sec:spectralSLAM}, we can view the
range-only SLAM problem as an instance of spectral state space
discovery. And, the Appendix (Sec.~\ref{sec:new-motion}) discusses how to
identify transition and measurement models given the learned states.
The same properties that make spectral methods appealing for system
identification carry over to our spectral SLAM algorithm:
computational efficiency, statistical consistency, and finite-sample
error bounds.

\subsection{Orthographic Structure From Motion}
In some ways the orthographic structure from motion (SfM) problem in
vision~\cite{Tomasi92} is very similar to the SLAM problem: the goal
is to recover scene geometry and camera rotations from a sequence of
images (compare with landmark geometry and robot poses from a sequence
of range observations).  And in fact, one popular solution for SfM is
very similar to the state space discovery step in spectral state space
identification.
The key idea in spectral SfM is that is that an image sequence can be
represented as a $2F\times P$ measurement matrix $W$, containing the
horizontal and vertical coordinates of $P$  points tracked through $F$
frames. If the images are the result of an orthographic camera
projection, then it is possible to show that $\text{rank}(W) = 3$. As
a consequence, the measurement matrix can be factored into the product
of two matrices $U$ and $V$, where $U$ contains the 3d positions of
the features and $V$ contains the camera axis
rotations~\cite{Tomasi92}. With respect to system identification, it
is possible to interpret the matrix $U$ as an observation model and
$V$ as an estimate of the system state.
Inspired by SfM, we reformulate range-only SLAM problem in a similar
way in Section~\ref{sec:spectralSLAM}, and then similarly solve the
problem with a spectral learning algorithm.  Also similar to SfM, we
examine the identifiability of our factorization, and give a
\emph{metric upgrade} procedure which extracts additional geometric
information beyond what the factorization gives us.

\subsection{Dimensionality-reduction-based Methods for Mapping}
Dimensionality reduction methods have recently provided an alternative to more traditional likelihood-based methods for mapping. In particular, the problem of finding a good map can be viewed as finding a (possibly nonlinear) \emph{embedding} of sensor data via methods like  multidimensional scaling (MDS) and manifold learning.

For example, MDS has been used to determine a Euclidean map of sensor locations where there is no distinction between landmark positions and robot positions~\cite{Shang2003}: instead \emph{all-to-all} range measurements are assumed for a set of landmarks. If some pairwise measurements are not available, these measurements can be approximated by some interpolation method, e.g. the geodesic distance between the landmarks~\cite{Tenenbaum00,Shang2003}.

Our problem differs from this previous work: in contrast to MDS, we have no landmark-to-landmark measurements and
 only inaccurate robot-to-robot measurements (from odometry, which may 
 not be present, and which often has significant errors when integrated
 over more than a short distance).  
Additionally, our smaller set of measurements 
introduces additional challenges not present in classical MDS: linear methods can recover the positions only up to a linear transformation.  This ambiguity 
forces changes compared to the MDS
algorithm: while MDS factors the all-to-all matrix of squared
 ranges, in Sec.~\ref{sec:landPos} we factor only a block of this matrix,
 then use either a metric upgrade step or a few global position
 measurements to resolve the ambiguity.  

A popular alternative to linear dimensionality reduction techniques like classical MDS is \emph{manifold learning}: nonlinearly mapping sensor inputs to a feature space that ``unfolds'' the manifold on which the data lies and \emph{then} applying dimensionality reduction. Such nonlinear dimensionality reduction has been used to learn maps of wi-fi networks and landmark locations when sensory data is thought to be nonlinearly related to the underlying Eucidean space in which the landmarks lie~\cite{Biggs05,Ferris2007,Yairi2007}. 
Unlike theses approaches, we show that \emph{linear} dimensionality reduction is sufficient to solve the range-only SLAM problem.
(In particular, \cite{Yairi2007} suggests solving range-only mapping using {nonlinear} dimensionality reduction. We not only show that this is unnecessary, but additionally show that linear dimensionality reduction is sufficient for localization as well.)
This greatly simplifies the learning algorithm and allows us to
provide strong statistical guarantees for the mapping portion of SLAM (Sec.~\ref{sec:consistency}).

\section{State Space  Discovery and Spectral SLAM}
\label{sec:spectralSLAM}
We start with SLAM from range data without odometry.
For now, we assume no noise, no missing
data, and batch processing.  We will generalize below:
Sec.~\ref{sec:orientation}
discusses how to recover robot orientation, 
Sec.~\ref{sec:spectral} discusses noise, and Sec.~\ref{sec:missing} discusses
missing
data and online SLAM\@. In the Appendix (Section~\ref{sec:new-motion}) we discuss learning motion and
measurement models.

\subsection{Range-only SLAM as Matrix Factorization}
\label{sec:landPos}
Consider the matrix $Y \in \mathbb{R}^{N \times T}$ of squared
ranges, with $N\geq 4$ landmarks and $T\geq 4$ time steps:
\begin{align}
Y & = \frac{1}{2}\left[\begin{array}{cccc}
d_{11}^2 & d_{12}^2 & \ldots & d_{1T}^2 \\
d_{21}^2 & d_{22}^2 & \ldots & d_{2T}^2 \\
\vdots & \vdots & \vdots & \vdots  \\
d_{N1}^2 & d_{N2}^2 & \ldots & d_{NT}^2 \\
\end{array}\right] 
\end{align}
where $d_{n,t}$ is the measured distance from the robot to landmark
$n$ at time step $t$.

The most basic version of our spectral SLAM method relies on the
insight that $Y$ \emph{factors} according to robot position $(x_t,y_t)$
and landmark position $(m_{n,x},m_{n,y})$. To see why, note
\begin{align}\label{eq:dist}
d_{n,t}^2 =  (m_{n,x}^2+m_{n,y}^2) -2m_{n,x}\cdot x_t -2m_{n,y}\cdot y_t+(x_t^2+y_t^2)
\end{align}
If we write $C_n= [(m_{n,x}^2+m_{n,y}^2)/2, m_{n,x}, m_{n,y}, 1]\tps$ and $X_t=
[1, -x_t, -y_t, (x_{t}^2+y_t^2)/2]\tps$, it is easy to see that 
$d_{n,t}^2=2C_n\tps X_t$.  So, $Y$ factors as $Y=CX$, where
$C\in\Re^{N\times 4}$ contains the
positions of landmarks,
\begin{align}
C = 
\left[ \begin{array}{cccc}
 (m_{1,x}^2+m_{1,y}^2)/2 & m_{1,x} & m_{1,y} &1 \\
 (m_{2,x}^2+m_{2,y}^2)/2 & m_{2,x} & m_{2,y} &1 \\
\vdots & \vdots & \vdots & \vdots\\
(m_{N,x}^2+m_{N,y}^2)/2& m_{N,x} & m_{N,y} & 1
\end{array} \right] 
\label{eq:C}
\end{align}
and $X \in \mathbb{R}^{4 \times T}$ contains the positions of the robot over time
\begin{align}
X   =  
\left[\begin{array}{ccc}
1 &  \ldots & 1 \\
-x_{1} & \ldots & -x_{T} \\
-y_{1} &  \ldots & -y_{T} \\
(x_1^2+y_1^2)/2  & \ldots & (x_T^2+y_T^2)/2 
\end{array}\right]
\label{eq:X}
\end{align}
If we can recover $C$ and $X$, we can read off the solution to the
SLAM problem. The fact that $Y$'s rank is at most 4 suggests that we
might be able to use a rank-revealing factorization of $Y$, such as
the singular value decomposition, to find $C$ and $X$.
Unfortunately, such a factorization only determines $C$ and $X$ up
to a linear transform: given an invertible matrix $S$, we can write $Y
= CX = CS^{-1}SX$. Therefore, factorization can only hope to recover
$U = CS^{-1}$ and $V = SX$.

\begin{figure}[!tb]
\begin{center}
\includegraphics[width=1\columnwidth]{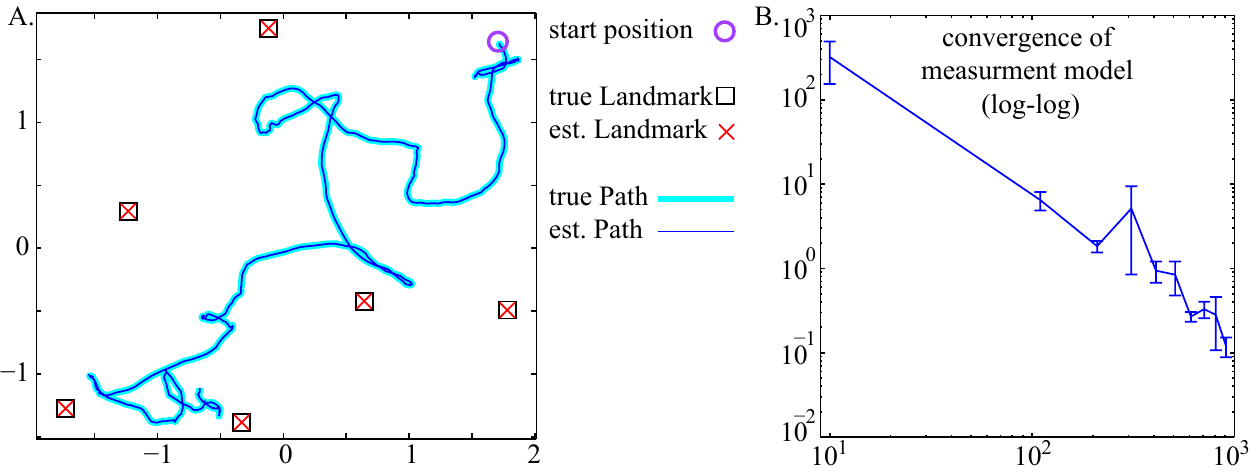}
\end{center}
\caption{Spectral SLAM on simulated data. See Section~\ref{sec:synthetic} 
  for details. A.) Randomly generated landmarks (6 of them) and
  robot path through the environment (500 timesteps). A SVD of the
  squared distance matrix recovers a linear transform of the landmark
  and robot positions. Given the coordinates of 4 landmarks, we can
  recover the landmark and robot positions in their original coordinates; or, since $500\geq
  9$, we can recover positions up to an orthogonal transform with no
  additional information. Despite noisy observations, the robot recovers
  the true path and landmark positions with very high accuracy.  B.)
  The convergence of the observation model $\widehat C_{5:6}$ for the
  remaining two landmarks: mean
  Frobenius-norm error vs.\ number of range readings received, averaged over 1000 randomly generated pairs of robot paths and environments.  Error bars indicate 95\% confidence intervals. } \label{fig:syntheticSLAM}
\end{figure}

To upgrade the factors $U$ and $V$ to a full metric map, we have two
options.  If global position estimates are available for at least four
landmarks, we can \emph{learn} the transform $S$ via
linear regression, and so recover the original $C$ and $X$.  
 This
 method works as long as we know at least four landmark positions.
Figure~\ref{fig:syntheticSLAM}A shows a simulated example.  

On the
other hand, if no global positions are known, the best we can hope to
do is recover landmark and robot positions up to an orthogonal
transform (translation, rotation, and reflection).  It turns out that
Eqs.~(\ref{eq:C}--\ref{eq:X}) provide enough additional geometric
constraints to do so:
in the Appendix (Sec.~\ref{sec:metric}) we show that, if we have at least
$9$
time steps and at least $9$ landmarks, and if each of
these point sets is non-singular in an appropriate sense, then we can
compute the metric upgrade in closed form. 
The idea is to fit a quadratic surface to the
rows of $U$,  
then change coordinates so that the
surface becomes the function in~(\ref{eq:C}).
 (By contrast, the usual metric upgrade for
orthographic structure from motion~\cite{Tomasi92}, which uses the
constraint that
camera projection matrices are orthogonal, requires a nonlinear
optimization.)

\subsection{SLAM with Headings}
\label{sec:orientation}

In addition to location, we often want the robot's global
heading $\theta$. 
We could get headings by post-processing our
learned positions,
but
in practice we can reduce variance by learning positions and headings
simultaneously.  We do so by adding
more features to our measurement matrix:
differences between successive pairs of
squared distances, scaled by velocity (which we can estimate
from odometry).  Since we need pairs of time steps, we now have $Y \in
\mathbb{R}^{2N \times T-1}$:

\begin{align}\label{eq:Y}
Y & = \frac{1}{2}\left[\begin{array}{cccc}
d_{11}^2 & d_{12}^2 & \ldots & d_{1T-1}^2 \\
\vdots & \vdots & \ddots & \vdots  \\
d_{N1}^2 & d_{N2}^2 & \ldots & d_{NT-1}^2 \\
\frac{d_{12}^2 - d_{11}^2}{v_1} & \frac{d_{13}^2 - d_{12}^2}{v_2} & \ldots &  \frac{d_{1T}^2 - d_{1T-1}^2}{v_{T-1}} \\
\vdots & \vdots & \ddots & \vdots  \\
\frac{d_{N2}^2-d_{N1}^2}{v_1} & \frac{d_{N2}^2-d_{N3}^2}{v_2} & \ldots & \frac{d_{NT}^2 - d_{NT-1}^2}{v_{T-1}} \\
\end{array}\right] 
\end{align}%
As before, we can factor $Y$ into  
a robot state matrix
and a landmark matrix.  The key new observation is that we can
write the new features in terms of $\cos(\theta)$ and $\sin(\theta)$:
\begin{align}
\label{eq:feat2}
\frac{d_{n,t+1}^2 -  d_{n,t}^2}{2v_t}  =&    - \frac{m_{n,x}(x_{t+1}  -  x_t)}{v_t}  - \frac{m_{n,y}(y_{t+1}  -  y_t)}{v_t}  + \frac{x_{t+1}^2  -  x_{t}^2  +  y_{t+1}^2  - y_t^2}{2v_t} \nonumber\\
=& - m_{n,x} \cos(\theta_t) - m_{n,y} \sin(\theta_t)  +\frac{x_{t+1}^2  -  x_{t}^2  +  y_{t+1}^2  -  y_t^2}{2v_t}
\end{align}
From Eq.~\ref{eq:dist} and Eq.~\ref{eq:feat2} it is easy to see that
$Y$ has rank at most 7 (exactly 7 if the robot path and landmark
positions are not singular): we have $Y=CX$, where $C \in \mathbb{R}^{N \times 7}$
contains functions of landmark positions and $X \in \mathbb{R}^{7
  \times T}$ contains functions of robot state,
  
\begin{align}
\label{eq:7dimstate}
C &=
\left[\begin{array}{ccccccc}
(m_{1,x}^2+m_{1,y}^2)/2 & m_{1,x} & m_{1,y} & 1 & 0 & 0 & 0 \\
\vdots & \vdots & \vdots & \vdots & \vdots & \vdots & \vdots \\
(m_{N,x}^2+m_{N,y}^2)/2 & m_{N,x} & m_{N,y} & 1 & 0 & 0 & 0\\
0 & 0& 0 & 0 &m_{1,x} & m_{1,y} & 1 \\
\vdots & \vdots & \vdots & \vdots & \vdots & \vdots & \vdots \\
0 &0 & 0 & 0 &m_{N,x} & m_{N,y} & 1
\end{array}\right]\\
X &=
\left[\begin{array}{ccc}
1 &  \ldots & 1 \\
-x_{1} & \ldots & -x_{T-1} \\
-y_{1}  & \ldots & -y_{T-1} \\
(x_1^2+y_1^2)/2 &  \ldots & (x_{T-1}^2+y_{T-1}^2)/2 \\
-\cos (\theta_1) &  \ldots & -\cos(\theta_{T-1})\\
-\sin(\theta_1) &  \ldots & -\sin(\theta_{T-1})\\
\frac{x_{2}^2 - x_{1}^2 + y_{2}^2 - y_1^2}{2v_1} & \hdots &\frac{x_{T}^2 - x_{T-1}^2 + y_{T}^2 - y_{T-1}^2}{2v_{T-1}}
\end{array}\right]
\end{align}%
As with the basic SLAM algorithm in Section~\ref{sec:landPos}, we can
factor $Y$ using SVD, this time keeping 7 singular values.  To make
the state space interpretable,
we can
then look at the top part of the learned transform of $C$: as long as
we have at least four landmarks in non-singular position, this block
will have exactly a three-dimensional nullspace (due to the three columns of zeros
in the top part of $C$).  After eliminating this nullspace, we can
proceed as before to learn $S$ and make the state space interpretable:
either use the coordinates of at least 4 landmarks as regression
targets, or perform a metric upgrade.  (See the Appendix,
Sec.~\ref{sec:metric}, for details).
 Once we have 
 positions, 
we can recover headings as angles between successive positions.

\subsection{A Spectral SLAM Algorithm}
\label{sec:spectral}

\begin{algorithm}[t] 
\caption{Spectral SLAM}
\label{alg:spectral}
\textbf{In}: \emph{i.i.d.}\ pairs of observations $\{ {o}_t,
{a}_t\}_{t=1}^T$; optional: measurement model for $\geq 4$ % (or more)
landmarks $C_{1:4}$ \\ % (by e.g.\ GPS) \\ 
\textbf{Out}:  measurement model (map) $\widehat C$, robot locations $\widehat X$ (the $t$th column is location at time $t$)\\
  \begin{algorithmic}[1]  \label{alg:robotSSID}
\STATE Collect observations and odometry into a matrix $\widehat Y$ (Eq.~\ref{eq:Y}) \vspace{2mm}
\STATE  Find the the top
$7$ singular values and vectors: $\langle \widehat U, \widehat \Lambda, \widehat
V^\top \rangle \leftarrow \text{SVD}(\widehat Y,7)$\\
The transformed measurement matrix is $\widehat CS^{-1} =
 \widehat U$ and robot states are $S\widehat X =  \widehat \Lambda
 \widehat V^\top$.\vspace{2mm}
 \STATE  Find $\widehat S$ via linear regression (from $\widehat U$ to
 $C_{1:4}$) or metric upgrade (see Appendix)\\ and return $\widehat C
 =\widehat U \widehat S$ and $\widehat X = \widehat S^{-1} \widehat \Lambda
 \widehat V^\top$

  \end{algorithmic}
  \end{algorithm}

The matrix factorizations of Secs.~\ref{sec:landPos}
and~\ref{sec:orientation} suggest a straightforward SLAM algorithm,
Alg.~\ref{alg:spectral}: 
build an empirical estimate $\widehat Y$ of $Y$ by sampling
observations as the robot traverses its environment, then apply a
rank-7 thin SVD, discarding the remaining singular values to
suppress noise.
\begin{align}
  \langle \widehat U, \widehat \Lambda, \widehat
V^\top \rangle \leftarrow \text{SVD}(\widehat Y,7)
\end{align}
Following Section~\ref{sec:orientation}, the left singular vectors
$\widehat U$ are an estimate of our transformed measurement matrix
$CS^{-1}$, and the weighted right singular vectors $ \widehat \Lambda
\widehat V^\top$ are an estimate of our transformed robot state $SX$.
We can then learn $S$ via regression or metric upgrade.

\paragraph{Statistical Consistency and Sample Complexity}

\label{sec:consistency}
Let $M \in \mathbb{R}^{N\times N}$ be the \emph{true} observation
covariance for a randomly sampled robot position, and let $\widehat M =
\frac{1}{T} \widehat Y\widehat Y^\top$ 
be the empirical covariance estimated from $T$ observations.  Then the
true and estimated measurement models are the top singular vectors of $M$
and $\widehat M$.
Assuming
that the noise in $\widehat M$ is zero-mean, as we include more data
in our averages, we will show below that the law of large numbers
guarantees that $\widehat M$ converges to the true covariance $M$.
So, our learning algorithm is
\emph{consistent} for estimating the range of $M$, i.e., the landmark locations.  (The estimated robot positions will typically not
converge, since we typically have a bounded effective number of observations
relevant to each robot position.  But, as we see each landmark again and
again, the robot position errors will average out, and we will recover
the true map.)

In more detail, we can give finite-sample bounds on the error in
recovering the true factors.  For simplicity of presentation we assume
that noise 
is i.i.d., although our algorithm will work for any zero-mean noise
process with a finite mixing time.  (The error bounds
will of course become weaker in proportion to mixing time, since we gain
less new information per observation.)
The argument (see the Appendix, Sec.~\ref{sec:complex}, for
details) has two pieces: standard concentration bounds 
show that each element of our estimated covariance approaches its
population value; then the continuity of the SVD shows that the
learned subspace also approaches its true value.  The final bound
is:
\begin{align}
|| \sin \Psi||_2 \leq  \frac{N c\sqrt{   \frac{2\log(T)} {T}}}{\gamma}
\end{align}
where $\Psi$ is the vector of canonical angles between the learned
subspace and the true one, $c$ is a constant depending on our error
distribution, 
and $\gamma$ is the true smallest nonzero eigenvalue of
the covariance.  In particular, this bound means that the sample
complexity is $\tilde O(\zeta^2)$ to achieve error $\zeta$.

\subsection{Extensions: Missing Data, Online SLAM, and System ID}

\paragraph{Missing data}
\label{sec:missing}
So far we have assumed that we receive range readings to all landmarks
at each time step. In practice this assumption is rarely satisfied: we
may receive range readings asynchronously, some range readings may be
missing entirely, and it is often the case that odometry data is
sampled faster than range readings.  Here we outline two methods for
overcoming this practical difficulty.

First, if a relatively small number of observations are missing, we
can use standard approaches for factorization with missing data. For
example, probabilistic PCA~\cite{Tipping99} estimates the missing
entries via an EM algorithm, and matrix completion~\cite{candes2009} uses a
trace-norm penalty to recover a low-rank factorization with high probability. However, for range-only data, often the fraction of missing data is high and the missing values are structural rather than random.

The second approach is interpolation:
we divide the data into overlapping subsets and then use local odometry information to interpolate the range data within each subset. To interpolate the data, we estimate a robot path by dead reckoning. For each point in the dead reckoning path we build the feature representation $[1, -x, -y, (x^2 + y^2)/2]^\top$. We then learn a linear model that predicts a squared range reading from these features (for the data points where range is available), as in Eq.~\ref{eq:dist}. Next we predict the squared range along the entire path.
 Finally we build the matrix $\widehat Y$ by averaging the locally
 interpolated range readings. This interpolation approach works much better
 in practice than
 the fully probabilistic approaches mentioned above, and was used in our experiments in Section~\ref{sec:results}.

\paragraph{Online Spectral SLAM}
\label{sec:online}
The algorithms developed in this section so far have had an important drawback:
unlike many SLAM algorithms, they are batch methods not online ones. The extension to online SLAM is straightforward: instead of first estimating $\widehat Y$ and then performing a SVD, we sequentially estimate our factors $\langle \widehat U, \widehat \Lambda, \widehat V^\top \rangle$ via online SVD~\cite{Brand2006,Boots-online-psr}.

\paragraph{Robot Filtering and System Identification}\label{sec:RSI}
So far, our algorithms have not directly used (or needed) a robot motion model in the learned state space.
However, an explicit motion model is required if we want to \emph{predict}
future sensor readings or \emph{plan} a course of action.  We have two
choices: we can derive a motion model from our
learned transformation $S$ between latent states and physical
locations,
or we can learn a motion
model directly from data using spectral system identification.
More details about both of these approaches can be found in the
Appendix, Sec.~\ref{sec:new-motion}.

\section{Experimental Results}
\label{sec:results}
We perform several SLAM and robot navigation experiments to illustrate and test the ideas proposed in this paper. 
First we show how our methods work in theory with synthetic experiments where complete observations are received at each point in time and \emph{i.i.d.}\ noise is sampled from a multivariate Gaussian distribution. Next we 
demonstrate our algorithm on data collected from a real-world robotic system with substantial amounts of missing data. 
Experiments were performed in Matlab, on a 2.66 GHz Intel Core i7 computer with 8 GB of RAM\@. In contrast to batch nonlinear optimization approaches to SLAM, the spectral learning methods described in this paper are \emph{very} fast, usually taking less than a second to run.

\subsection{Synthetic Experiments}
\label{sec:synthetic}
Our simulator randomly places 6 landmarks in a 2-D
environment. A simulated robot then randomly moves through the environment for
500 time steps and receives a range reading to each one of the
landmarks at each time step. The range readings are perturbed by noise
sampled from a Gaussian distribution with variance equal to 1\% of the
range. Given this data, we apply the algorithm from
Section~\ref{sec:spectral} to solve the SLAM problem. We use the
coordinates of 4 landmarks to learn the linear transform $S$ and
recover the true state space, as shown in
Figure~\ref{fig:syntheticSLAM}A. The results indicate that we can
accurately recover both the landmark locations and the robot path.

We also investigated the empirical convergence rate of our observation
model (and therefore the map) as the number of range readings
increased. To do so, we generated 1000 different random pairs of
environments and robot paths. For each pair, we repeatedly performed
our spectral SLAM algorithm on increasingly large numbers of range
readings and looked at the difference between our estimated
measurement model (the robot's map) and the true measurement model,
excluding the landmarks that we used for reconstruction: $\|
\widehat C_{5:6} - C_{5:6}\|_\mathcal{F}$. The results are shown in
Figure~\ref{fig:syntheticSLAM}B, and show that our estimates steadily
converge to the true model, corroborating our theoretical results
(in Section~\ref{sec:consistency} and the Appendix).

\subsection{Robotic Experiments}

\begin{figure*}[!tb]
\begin{center}
\includegraphics[width=1\columnwidth]{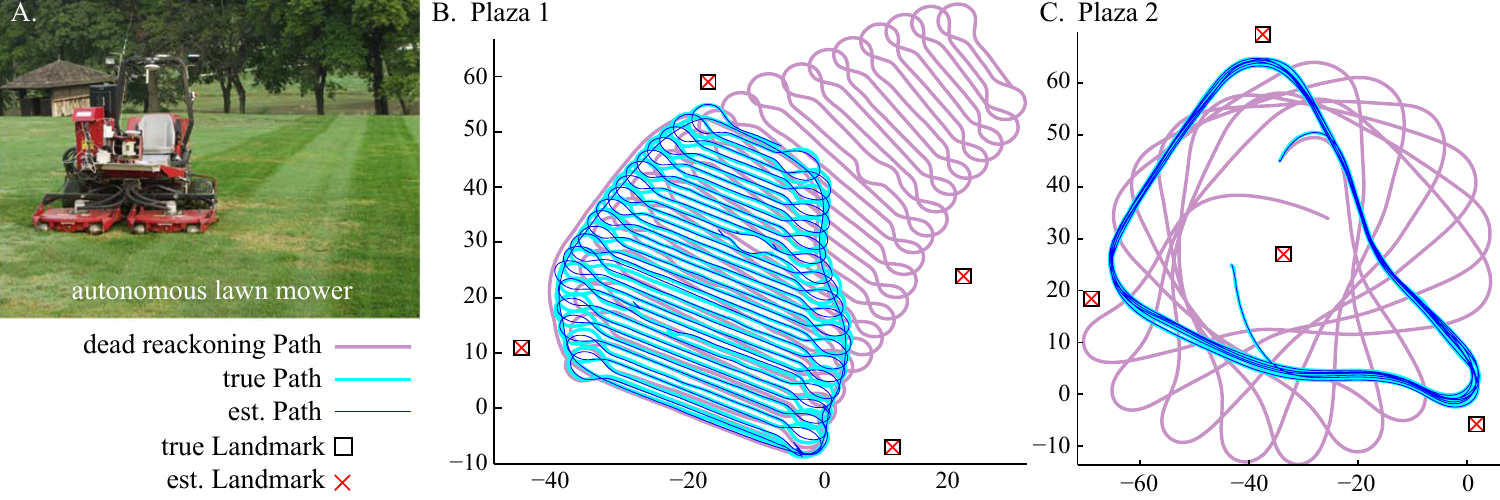}
\end{center}
\caption{The autonomous lawn mower and spectral SLAM\@. A.) The robotic lawn mower platform. 
  B.) In the first experiment, the robot traveled 1.9km receiving 3,529 range measurements. This path minimizes the effect of heading error by balancing the number of left turns with an equal number of right turns in the robot's odometry (a commonly used path pattern in lawn mowing applications).  The light blue path indicates the robot's true path in the environment, light purple indicates dead-reckoning path, and dark blue indicates the spectral SLAM localization result.
   C.) In the second experiment, the robot traveled 1.3km receiving 1,816 range measurements.  This path highlights the effect of heading error on dead reckoning performance by turning in the same direction repeatedly.  Again, spectral SLAM is able to accurately recover the robot's path.
} \label{fig:lawn}
\end{figure*}
We used two freely available range-only SLAM data sets collected from an autonomous lawn mowing robot~\cite{Djugash2010}, shown in Fig.~\ref{fig:lawn}A.%
\footnote{http://www.frc.ri.cmu.edu/projects/emergencyresponse/RangeData/index.html}
These ``Plaza'' datasets were collected via radio nodes from
Multispectral Solutions that use time-of-flight of ultra-wide-band
signals to provide inter-node ranging measurements. (Additional details on the experimental setup can be found in~\cite{Djugash2010}.)  
 This system
produces a time-stamped range estimate between the mobile robot and
stationary nodes (landmarks) in the environment. The landmark radio
nodes are placed atop traffic cones approximately 138cm above the
ground throughout the environment, and one node was placed on top of
the center of the robot's coordinate frame (also 138cm above the
ground). The robot odometry 
(dead reckoning) comes from an onboard fiberoptic gyro and wheel
encoders. The
two environmental setups, including the locations of the landmarks, the
dead reckoning paths, and the ground truth paths, are shown in
Figure~\ref{fig:lawn}B-C\@.  The ground truth paths have 2cm accuracy
according to~\cite{Djugash2010}.  

The two Plaza datasets that we used to evaluate our algorithm have very different characteristics.
 In ``Plaza 1,'' the robot travelled 1.9km, occupied 9,658 distinct poses, and received 3,529 range measurements. The path taken is a typical lawn mowing pattern that balances left turns with an equal number of right turns; this type of pattern minimizes the effect of heading error. In ``Plaza 2,'' the robot travelled 1.3km, occupied 4,091 poses, and received 1,816 range measurements. The path taken is a loop which amplifies the effect of heading error. The two data sets were both very sparse, with approximately 11 time steps
(and up to 500 steps) between range readings for the worst landmark.
We first interpolated the missing range
readings with the method of
Section~\ref{sec:missing}. Then we applied the rank-7 spectral SLAM
algorithm of Section~\ref{sec:spectral}; the results are
depicted in Figure~\ref{fig:lawn}B-C. Qualitatively, we see that the
robot's localization path conforms to the true path.

\begin{figure}[!tb]
 \begin{minipage}[c]{0.72\textwidth}
\begin{tabular}{|l||r|r|} 
\hline
{\bf Method} & \bf Plaza 1& \bf Plaza 2 \\
\hline\hline
 Dead Reckoning (full path) & 15.92m & 27.28m \\
\hline\hline
 Cartesian EKF (last, best 10\%) & 0.94m & 0.92m \\   
\hline

FastSLAM (last, best 10\%)& 0.73m &1.14m \\
\hline
ROP EKF (last, best 10\%) & { 0.65m}& 0.87m \\
\hline\hline
Batch Opt. (worst 10\%) & 1.04m & { 0.45m}  \\
\hline
Batch Opt. (last 10\%) & 1.01m & { 0.45m} \\
\hline
Batch Opt. (best 10\%) & 0.56m & 0.20m   \\
\hline
Batch Opt. (full path) & 0.79m  & 0.33m \\
\hline\hline
Spectral SLAM (worst 10\%) & {1.01m}& { 0.51m} \\
\hline
Spectral SLAM (last 10\%) &0.98m&0.51m \\
\hline
Spectral SLAM (best 10\%) & {  0.59m}& { 0.22m} \\
\hline
Spectral SLAM (full path) & { 0.79m}& { 0.35m} \\
\hline\hline
Spectral + Batch Optimization (worst 10\%) & {\bf 0.89m} & {\bf 0.40m}\\
\hline
Spectral + Batch Optimization (last 10\%) &{ 0.81m} & {\bf 0.32m} \\
\hline
Spectral + Batch Optimization (best 10\%) & {\bf  0.54m}& {\bf 0.18m}\\
\hline
Spectral + Batch Optimization (full path) & {\bf 0.69m}& {\bf 0.30m} \\
\hline
\end{tabular}
\end{minipage}
 \begin{minipage}[c]{.27\textwidth}
\includegraphics[width=1\columnwidth]{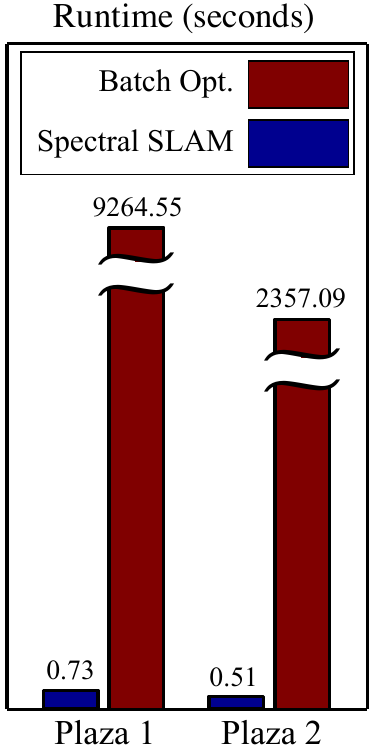}
\end{minipage}
\caption{Comparison of Range-Only SLAM Algorithms. The table shows
  Localization RMSE.  Spectral SLAM has localization accuracy
  comparable to batch optimization on its own. The best results
  (boldface entries) are obtained by initializing nonlinear batch
  optimization with the spectral SLAM solution. The graph compares
  runtime of Gauss-Newton batch optimization with spectral
  SLAM\@. Empirically, spectral SLAM is 3-4 \emph{orders of magnitude}
  faster than batch optimization on the autonomous lawnmower
  datasets. }
\label{tab:results}
\end{figure}

In addition to the qualitative results, we quantitatively compared
spectral SLAM to a number of different competing range-only SLAM
algorithms. The localization root mean squared error (RMSE) in meters
for each algorithm is shown in Figure~\ref{tab:results}.  The baseline
is 
dead reckoning (using only the robot's odometry information).  Next are several standard online range-only SLAM algorithms, summarized in~\cite{Djugash2010}. These algorithms included the Cartesian EKF,  FastSLAM~\cite{Montemerlo02} with 5,000 particles, and the ROP-EKF~\cite{Djugash2008}. These previous
results only reported the RMSE for the last $10\%$ of the path,
which is typically the \emph{best} $10\%$ of the path (since it gives
the most time to recover from initialization problems). The full path
localization error
can be considerably worse, particularly for the initial portion of the
path---see Fig.~5 (right) of~\cite{Djugash2008}.

We also compared to batch nonlinear optimization, via Gauss-Newton as
implemented in Matlab's {\tt{fminunc}} (see~\cite{Kehagias2006} for
details). This approach to solving the range-only SLAM problem can be
very data efficient, but is subject to local optima and is very
computationally intensive. We followed the suggestions
of~\cite{Kehagias2006} and initialized with the dead-reckoning
estimate of the robot's path. The 
algorithm took roughly 2.5 hours to converge on Plaza 1, and 45
minutes to converge on Plaza 2. Under most evaluation metrics, the nonlinear batch
algorithm handily beats the EKF-based alternatives.
 
Finally, we ran our spectral SLAM algorithm on the same data sets. In
contrast to Gauss-Newton, spectral SLAM is \emph{statistically
  consistent}, and much faster: the bulk of the computation is the
fixed-rank SVD, so the time complexity of the algorithm is $O((2N)^2 T)$ where $N$ is the number of landmarks and $T$ is the number of time steps. Empirically, spectral SLAM produced results that were comparable to batch optimization in  3-4 \emph{orders of magnitude} less time (see Figure~\ref{tab:results}). 

Spectral SLAM can also be used as an initialization procedure for nonlinear batch optimization. This strategy combines the best of both algorithms by allowing the locally optimal nonlinear optimization procedure to start from a theoretically guaranteed good starting point. Therefore, the local optimum found by nonlinear batch optimization should  be \emph{no worse} than the spectral SLAM solution and likely much better than the batch optimization seeded by dead-reckoning. Empirically, we found this to be the case (Figure~\ref{tab:results}). If time and computational resources are scarce, then we believe that spectral SLAM is clearly the best approach; if computation is not an issue, the best results will almost certainly be found by refining the spectral SLAM solution using a nonlinear batch optimization procedure.

\section{Conclusion}
\label{sec:conclusion}
We proposed a novel solution for the range-only SLAM problem that differs substantially from previous approaches. The essence of this new approach is to formulate SLAM as a factorization problem, which allows us to derive a local-minimum free spectral learning method that is closely related to SfM and spectral approaches to system identification. We provide theoretical guarantees for our algorithm, discuss how to derive an online algorithm, and show how to generalize to a full robot system identification algorithm. Finally, we demonstrate that our spectral approach to SLAM beats other state-of-the-art SLAM approaches on real-world range-only SLAM problems.

\section*{Acknowledgements}
Byron Boots and Geoffrey Gordon were supported by ONR MURI grant
number N00014-09-1-1052.  Byron Boots was supported by the NSF under
grant number EEEC-0540865.

\clearpage

\bibliographystyle{mlapa2}
\bibliography{references}  

\clearpage

\section{Appendix}
\subsection{Metric Upgrade for Learned Map}
\label{sec:metric}
In the main body of the paper, we assumed that global position estimates of at least four landmarks were known. When these landmarks are known, we can recover all of the estimated landmark positions and robot locations.

In many cases, however, no global positions are known; the best we can
hope to do is recover landmark and robot positions up to an orthogonal
transform (translation, rotation, and reflection).  It turns out that
Eqs.~(\ref{eq:C}--\ref{eq:X}) provide enough geometric constraints to
perform this metric upgrade, as long as we have at least $9$
landmarks and at least $9$ time steps, and as long as $C$ and $X$ are
\emph{nonsingular} in the following sense: define the matrix $C_2$,
with the same number of rows as $C$ but 10 columns, whose $i$th row
has elements $c_{i,j}c_{i,k}$ for $1\leq j\leq k\leq 4$ (in any fixed
order).  Note that the rank of $C_2$ can be at most 9: from
Eq.~\ref{eq:C}, we know that
$c_{i,2}^2+c_{i,3}^2-2c_{i,4} = 0$, and each of the three terms in
this function is a multiple of a column of $C_2$.  We will say that
$C$ is nonsingular if $C_2$ has rank exactly 9, i.e., is rank
deficient by exactly 1 dimension.  The conditions for $X$ are
analogous, swapping rows for columns.\footnote{For intuition, a set of
  landmarks or robot positions that all lie on the same quadratic
  surface (line, circle, parabola, etc.)  will be singular.  Some
  higher-order constraints will also lead to singularity; e.g., a set
  of points will be singular if they all satisfy
  $\frac{1}{2}(x_i^2+y_i^2)x_i + y_i=0$, since each of the two terms
  in this function is a column of $C_2$.}

To derive the metric upgrade, suppose that we start from an $N\times
4$ matrix $U$ of learned landmark coordinates and an $4\times N$
matrix $V$ of learned robot coordinates from the algorithm of
Sec.~\ref{sec:landPos}.
And, suppose that we have at least 9 nonsingular landmarks and robot
positions.
  We would like to transform the learned
coordinates into two new matrices $C$ and $X$ such that 
\begin{align}
c_1 &\approx 1 \label{eq:u1}\\
c_4 &\approx \frac{1}{2}c_2^2 + \frac{1}{2}c_3^2 \label{eq:u2}\\
x_4 &\approx 1 \label{eq:v1}\\
x_1 &\approx \frac{1}{2}x_2^2 + \frac{1}{2}x_3^2 \label{eq:v2}
\end{align}
where $c$ is a row of $C$ and $x$ is a column of $X$.

At a high level, we first fit a quadratic surface to the rows of $U$,
then transform this surface so that it satisfies
Eq.~\ref{eq:u1}--\ref{eq:u2}, and scale the surface so that it
satisfies Eq.~\ref{eq:v1}.  Our surface will then automatically also
satisfy Eq.~\ref{eq:v2}, since $X$ must be metrically correct if $C$ is.

In more detail, we first (step i) linearly transform each row of $U$
into approximately the form $(1, r_{i,1}, r_{i,2}, r_{i,3})$: we use
linear regression to find a coefficient vector $a\in\Re^4$ such that
$Ua\approx \bf 1$, then set $R = UQ$ where $Q\in\Re^{4\times 3}$ is an
orthonormal basis for the nullspace of $a\tps$.  After this step, our
factorization is $(U T_1)(T_1^{-1}V)$, where $T_1=\left(a\ Q\right)$.

Next (step ii) we fit an implicit quadratic surface to the rows of $R$
by finding $10$ coefficients $b_{jk}$ (for $0\leq j\leq k\leq 3$) such
that
\begin{align*}
0\ \approx\,\ &
b_{00} +
b_{01} r_{i,1} + 
b_{02} r_{i,2} + 
b_{03} r_{i,3} + {}\\
&b_{11} r_{i,1}^2 +
b_{12} r_{i,1} r_{i,2} + 
b_{13} r_{i,1} r_{i,3} + 
b_{22} r_{i,2}^2 + 
b_{23} r_{i,2} r_{i,3} + 
b_{33} r_{i,3}^2
\end{align*}
To do so, we form a matrix $S$ that has the same number of rows as $U$
but 10 columns.  The elements of row $i$ of $S$ are $r_{i,j}r_{i,k}$
for $0\leq j\leq k\leq 3$ (in any fixed order).  Here, for convenience, we
define $r_{i,0}=1$ for all $i$.  Then we find a vector $b\in\Re^{10}$ that
is approximately in the nullspace of $S\tps$ by taking a singular
value decomposition of $S$ and selecting the right singular vector
corresponding to the smallest singular value.  Using this vector, we
can define our quadratic as $0 \approx \frac{1}{2}r\tps H r + \ell\tps
r + b_{00}$, where $r$ is a row of $R$, and the Hessian matrix $H$ and
linear part $\ell$ are given by:
\[
H = \left(
\begin{array}{ccc}
\frac{1}{2} b_{11} & b_{12} & b_{13}\\
b_{21} & \frac{1}{2} b_{22} & b_{23}\\
b_{31} & b_{32} & \frac{1}{2} b_{33}
\end{array}
\right)
\qquad
\ell = \left(
\begin{array}{c}
b_{01}\\
b_{02}\\
b_{03}
\end{array}
\right)
\]
Over the next few steps we will transform the coordinates in $R$ to
bring our quadratic into the form of Eq.~\ref{eq:u2}: that is, one
coordinate will be a quadratic function of the other two, there will
be no linear or constant terms, and the quadratic part will be
spherical with coefficient $\frac{1}{2}$.

We start (step iii) by transforming coordinates so that our quadratic
has no cross-terms, i.e., so that its Hessian matrix is diagonal.
Using a $3\times 3$ singular value decomposition, we can factor
$H=MH'M\tps$ so that $M$ is orthonormal and $H'$ is diagonal.  If we
set $R' = RM$ and $\ell' = M\ell$, and write $r'$ for a row of
$R'$, we can equivalently write our quadratic as $0 =
\frac{1}{2}(r')\tps H' r' + (\ell')\tps r' + b_{00}$, which has a
diagonal Hessian as desired.  After this step, our factorization is
$(U T_1T_2)(T_2^{-1}T_1^{-1}V)$, where
\[ T_2=\left(\begin{array}{cc} 1 & 0 \\ 0 & M
\end{array}\right)
\]
Our next step (step iv) is to turn our implicit quadratic surface into
an explicit quadratic function.  For this purpose we pick one of the
coordinates of $R'$ and write it as a function of the other two.  In
order to do so, we must have zero as the corresponding diagonal
element of the Hessian $H'$---else we cannot guarantee that we can
solve for a unique value of the chosen coordinate.  So, we will take
the index $j$ such that $H'_{jj}$ is minimal, and set $H'_{jj}=0$.
Suppose that we pick the last coordinate, $j=3$.  (We can always
reorder columns to make this true; SVD software will typically do so
automatically.)  Then our quadratic becomes
\begin{align*}
0 &= \frac{1}{2}H'_{11} (r_1')^2 +
\frac{1}{2}H'_{22} (r_2')^2 + \ell_1' r_1' + \ell_2' r_2' + \ell_3'
r_3' + b_{00}\\
r_3' &= -\frac{1}{\ell_3'}\left[
\frac{1}{2}H'_{11} (r_1')^2 +
\frac{1}{2}H'_{22} (r_2')^2 + \ell_1' r_1' + \ell_2'
r_2' + b_{00}
\right]
\end{align*}
Now (step v) we can shift and rescale our coordinates one more time to
get our quadratic in the desired form: translate so that the linear
and constant coefficients are $0$, and rescale so that the quadratic
coefficients are $\frac{1}{2}$.  For the translation, we define new
coordinates $r''=r'+c$ for $c\in\Re^3$, so that our quadratic becomes
\begin{align*}
r_3'' &= c_3 - \frac{1}{\ell_3'}\left[
\frac{1}{2}H'_{11} (r_1''-c_1)^2 +
\frac{1}{2}H'_{22} (r_2''-c_2)^2 + \ell_1' (r_1''-c_1) + \ell_2'
(r_2''-c_2) + b_{00}
\right]
\end{align*}
By expanding and matching coefficients, we know $c$
must satisfy
\begin{align*}
0&=\frac{H'_{11}}{\ell_3'}c_1
-\frac{\ell_1'}{\ell_3'}&\text{(coefficient of }r''_1\text{)}\\
0&=\frac{H'_{22}}{\ell_3'}c_2 -\frac{\ell_2'}{\ell_3'}
&\text{(coefficient of }r''_2\text{)}\\
0&=c_3 - \frac{H'_{11}}{2\ell_3'}c_1^2 -\frac{H'_{22}}{2\ell_3'}c_2^2 +
\frac{\ell_1'}{\ell_3'}c_1 + \frac{\ell_2'}{\ell_3'}c_2 -
b_{00}/\ell'_3
&\text{(constant)}
\end{align*}
The first two equations are linear in $c_1$ and $c_2$ (and don't
contain $c_3$).  So, we can solve directly for $c_1$ and $c_2$; then
we can plug their values into the last equation  to find $c_3$.
For the scaling, the coefficient of $r_1''$ is now
$-\frac{H'_{11}}{2\ell_3'}$, and that of $r_2''$ is now
$-\frac{H'_{22}}{2\ell_3'}$.  So, we can just scale these two
coordinates separately to bring their coefficients to $\frac{1}{2}$.

After this step, our factorization is $U'V'$, where $U'=U
T_1T_2T_3$ and $V'=T_3^{-1}T_2^{-1}T_1^{-1}V$, and
\[ T_3=
\left(\begin{array}{cccc}
1 & 0 & 0 & 0\\
c_1 & -\frac{\ell_3'}{H'_{11}} & 0 & 0\\
c_2 & 0 & -\frac{\ell_3'}{H'_{22}}& 0\\
c_3 & 0 & 0 & 1
\end{array}\right)
\]
The left factor $U'$ will now satisfy Eq.~\ref{eq:u1}--\ref{eq:u2}.
We still have one last useful degree of freedom: if we set $C=U'T_4$,
where
\[
T_4 = 
\left(\begin{array}{cccc}
1 & 0 & 0 & 0\\
0 & \mu & 0 & 0\\
0 & 0 & \mu & 0\\
0 & 0 & 0 & \mu^2
\end{array}\right)
\]
for any $\mu\in\Re$, then $C$ will still satisfy
Eq.~\ref{eq:u1}--\ref{eq:u2}.  So (step vi), we will pick $\mu$ to
satisfy Eq.~\ref{eq:v1}: in particular, we set $\mu =
\sqrt{\text{mean}(V'_{4,:})}$, so that when we set $X=T_4^{-1}V'$,
the last row of $X$ will have mean 1.

If we have 7 learned coordinates in $U$ as in
Sec.~\ref{sec:orientation}, we need to find a subspace of 4
coordinates in order to perform metric upgrade.  To do so, we take
advantage of the special form of the correct answer, given in
Eq.~\ref{eq:7dimstate}: in the upper block of $C$ in
Eq.~\ref{eq:7dimstate}, three coordinates are identically zero.  Since
$U$ is a linear transformation of $C$, there will be three linear
functions of the top block of $U$ that are identically zero (or
approximately zero in the presence of noise).  As long as the landmark
positions are nonsingular, we can use SVD on the
top block of $U$ to find and remove these linear functions (by setting
the smallest three singular values to zero), then proceed as above
with the four remaining coordinates.

\subsection{Sample Complexity for the Measurement Model (Robot Map)}
\label{sec:complex}
Here we provide the details on how our estimation error scales with
the number $T$ of training examples---that is, the scaling of
the difference between the estimated measurement model $\widehat U$,
which contains the location of the landmarks, and its population
counterpart.

Our bound has two parts. First we use a standard concentration bound
(the Azuma-Hoeffding inequality) to show that each element of our estimated covariance $\widehat M = \widehat Y\widehat Y^\top$ approaches its population value. 
We start by rewriting the empirical covariance matrix as a vector
summed over multiple samples:
\begin{align*} 
\text{vec}\left (\widehat M \right ) = \frac{1}{T}\sum_{t=1}^T \Upsilon_{:,t}
\end{align*}
where $\Upsilon = (\widehat Y \odot \widehat Y)^\top$ is the matrix of column-wise Kronecker products of the observations $\widehat Y$. 
We assume that each element of $\Upsilon$ minus its expectation
$\EE\Upsilon_i$ is bounded by a constant $c$; we can derive $c$ from
bounds on anticipated errors in distance measurements and odometry
measurements.
\begin{align*}
| \Upsilon_{i,t} - \EE\Upsilon_{i} | \leq c, \quad \forall_{i,t}
\end{align*}
Then the Azuma-Hoeffding inequality bounds the probability that the
empirical sum differs too much from its population value: for any
$\alpha\geq 0$ and any $i$,
\begin{align*}
\PP \left [\left|\sum_{t = 1}^T  (\Upsilon_{i,t} -
    \EE\Upsilon_{i})\right| \geq \alpha \right ] \leq 2 e^{-\alpha^2/2Tc^2}
\end{align*}
If we pick $\alpha = \sqrt{2Tc^2 \log(T)}$, then we can rewrite the probability in terms of $T$:
\begin{align*}
\PP \left [ \frac{1}{T}\left|\sum_{t = 1}^T  (\Upsilon_{i,t} - \EE\Upsilon_{i}) \right|  \geq c\sqrt{\frac{2\log(T)}{T}} \right ] \leq 2e^{-\log(T)}
\end{align*}
which means that the probability decreases as $O(\frac{1}{T})$ and
the threshold decreases as $\tilde O(\frac{1}{\sqrt{T}})$.

We can then use a union bound over all $(2N)^2$ covariance elements (since $\widehat Y \in \mathbb{R}^{2N \times T}$):
\begin{align*}
\PP \left [ \forall i\, \left|\frac{1}{T}\sum_{t = 1}^T  \Upsilon_{i,t} - \EE\Upsilon_{i}\right|  \geq c \sqrt{\frac{2\log(T)}{T}} \right ] \leq  8 N^2/T
\end{align*}
That is, with
high probability, the entire empirical covariance matrix $\widehat M$
will be close (in max-norm) to its expectation.%\\

Next we use the continuity of the SVD to show that the learned
subspace approaches its true value.
Let $\widehat M = M + E$, where $E$ is the perturbation (so the
largest element of $E$ is bounded). 
  Let $\widehat U$ be the output of
SVD, and let $U$ be the population value (the top singular vectors
of the true $M$).
Let $\Psi$ be the matrix of
canonical angles between $\text{range}(U)$ and $\text{range}(\widehat
U)$. Since we know the exact rank of the true $M$ (either 4 or 7),
the last (4th or 7th) singular value of $M$ will be positive; call it
$\gamma>0$.  So, by Theorem 4.4 of Stewart and Sun~\cite{stewart-sun:1990},
\begin{align*}
|| \sin \Psi||_2 \leq \frac{||E||_2}{\gamma}
\end{align*}
This result uses a 2-norm bound on $E$, but the bound
we showed above is in terms of the largest element of $E$.
But, the 2-norm can be bounded in terms of the largest element:
\begin{align*}
|| E||_2 \leq N \max_{ij}|E_{ij}|
\end{align*}
Finally, the result is that we can bound the canonical angle:  
\begin{align*}
|| \sin \Psi||_2 \leq  \frac{N c\sqrt{   \frac{2\log(T)} {T}}}{\gamma}
\end{align*}
In other words, the canonical angle shrinks at a rate of
$\tilde O(\frac{1}{\sqrt{T}})$, with probability at least $1-\frac{8N^2}{T}$.
 
\subsection{The Robot as a Nonlinear Dynamical System}
\label{sec:new-motion}
Once we have learned an interpretable state space via the algorithm of
Section~\ref{sec:spectral}, we can simply write down the nominal robot
dynamics in this space.  The accuracy of the resulting model will
depend on how well our sensors and actuators follow the nominal
dynamics, as well as how well we have learned the transformation $S$
to the interpretable version of the state space.

In more detail, we model the robot as a controlled nonlinear dynamical
system.  The evolution is governed by the
following state space equations, which generalize~(\ref{eq:kalman-motion-measurement}):
 \begin{align}
 s_{t+1} &= f(s_t, a_t) + \epsilon_t \\
 o_t &= h(s_t) + \nu_t
 \end{align}
Here $s_t \in \mathbb{R}^{k}$ denotes the hidden state, $a_t \in
\mathbb{R}^{l}$ denotes the control signal, $o_t \in \mathbb{R}^{m}$
denotes the observation, $\epsilon_t \in \mathbb{R}^k$ denotes the
state noise, and $\nu_t \in \mathbb{R}^{m}$ denotes the observation
noise. For our range-only system, following the decomposition of
Section~\ref{sec:spectralSLAM}, we have:
 
\begin{align}\label{eq:settings}
s_{t} = \left[\begin{array}{c}
1 \\
-x_{t} \\
-y_{t}\\
(x_t^2+y_t^2)/2  \\
-\cos (\theta_t) \\
-\sin(\theta_t) \\
\frac{x_{t+1}^2 - x_{t}^2 + y_{t+1}^2 - y_t^2}{2v_t} 
\end{array}\right]
,\hspace{.5mm}
o_{t} = \left[\begin{array}{c}
d_{1t}^2/2  \\
\vdots  \\
d_{Nt}^2/2  \\
\frac{{d_{1t+1}^2 - d_{1t}^2}}{2v_t}  \\
\vdots   \\
\frac{d_{Nt+1}^2-d_{Nt}^2}{2v_t}\\
\end{array}\right]
,\hspace{.5mm}
a_{t} = \left[\begin{array}{c}
v_t \\
\cos(\omega_t)\\
\sin(\omega_t)\\
\end{array}\right]
\end{align}

\noindent Here $v_t$ and $\omega_t$ are the translation and rotation calculated from the robot's odometry. 
A nice property of this model is that expected observations are a \emph{linear} function of state:
\begin{align}
\label{eq:linobsmod}
 h(s_t) &= Cs_t
\end{align}

The dynamics, however, are \emph{nonlinear}: see
Eq.~\ref{eq:dynamics}, which can easily be derived from the basic
kinematic motion model for a wheeled robot~\cite{Thrun2005}.

 \begin{align}\label{eq:dynamics}
f(s_t, a_t) &= \left [ \begin{array}{c} 1 \\
-x_t -v_t \cos(\theta_t) \\
-y_t -v_t \sin(\theta_t) \\
\frac{x_t^2 + y_t^2}{2} +  v_{t} x_t \cos(\theta_t) + v_{t} y_t \sin(\theta_t)  + \frac{v^2_{t} \cos^2(\theta_t)+ v^2_{t} \sin^2(\theta_t)}{2}\\
- \cos(\theta_t)\cos(\omega_{t}) + \sin(\theta_t)\sin(\omega_{t})  \\
- \sin(\theta_t)\cos(\omega_{t}) + \cos(\theta_t)\sin(\omega_{t}) \\
\text{$[x_t\cos(\theta_t)\cos(\omega_t) - x_t \sin(\theta_t)\sin(\omega_t) + v_t\cos^2(\theta_t)\cos(\omega_t) +{}$}\\
\text{$y_t\sin(\theta_t)\cos(\omega_t)- y_t\sin(\omega_t)\cos(\theta_t) + v_t\sin^2(\theta_t)\cos(\omega_t) - {}$}\\
\text{$2v_t\cos(\theta_t)\sin(\theta_t)\sin(\omega_t)]$}
\end{array} \right ]
 \end{align}

\subsubsection{Robot System Identification}\label{sec:RSID}
To apply the model of Section~\ref{sec:new-motion}, it is essential
that we maintain states in the physical coordinate frame, and not just
the linearly transformed coordinate frame---i.e., $\widehat C$ and not
$\widehat U = \widehat CS^{-1}$.  So, to use this model, we must first
learn $S$ either by regression or by metric upgrade.

However, it is possible instead to use
\emph{system identification} to learn to filter directly in the raw
state space $\widehat U$.  We conjecture that it may be more robust to
do so, since we will not be sensitive to errors in the metric upgrade
process (errors in learning $S$), and since we can learn to compensate
for some deviations from the nominal model of
Section~\ref{sec:new-motion}.

To derive our system identification algorithm, we can explicitly
rewrite $f(s_t,a_t)$ as a nonlinear feature-expansion map followed by
a linear projection. Our algorithm will then just be to use linear
regression to learn the linear part of $f$.

First, let's look at the dynamics for the special case of
$S = I$.  Each additive term in Eq.~\ref{eq:dynamics} is the product
of at most two terms in $s_t$ and at most two terms in $a_t$. Therefore,
we define %a nonlinear mapping 
$\phi(s_t, a_t) :=  s_t \otimes s_t \otimes \bar a_t \otimes \bar a_t$,
where $\bar a_t = [ 1, a_t ]\tps$ and $\otimes$ is the Kronecker product. (Many of the dimensions of
$\phi(s_t, a_t)$ are duplicates; for efficiency we would delete these
duplicates, but for simplicity of notation we keep them.)
Each additive term in Eq.~\ref{eq:dynamics} is a multiple of an element of $\phi(s_t, a_t)$, so we can write the dynamics as:
\begin{align}
\label{eq:philintrans}
s_{t+1} = N\phi(s_t,a_t) + \epsilon_t
\end{align}
where $N$ is a linear function that picks out the correct entries to form Eq.~\ref{eq:dynamics}. 

Now, given an invertible matrix $S$, we can rewrite $f(s_t,a_t)$ as an \emph{equivalent} function in the transformed state space:
\begin{align}
Ss_{t+1} &= \bar f(Ss_t,a_t) + S\epsilon_t
\end{align}
To do so, we use the identity $(Ax)\otimes(By)=(A\otimes B)(x\otimes
y)$.  Repeated application yields
\begin{align}
\phi(Ss_t,a_t) &= Ss_t \otimes Ss_t \otimes \bar a_t \otimes \bar a_t\nonumber \\
&= (S\otimes S \otimes I \otimes I) (s_t \otimes s_t\otimes \bar a_t
\otimes \bar a_t )\nonumber\\
&= \bar S\,\phi(s_t,a_t)
\end{align}
where $\bar S = S\otimes S \otimes I \otimes I$.  Note that
$\bar S$ is invertible (since $\text{rank}(A \otimes B) =
\text{rank}(A)\,\text{rank}(B)$); so, we can write
\begin{align}
\bar f(Ss_t, a_t) &= SN\bar S^{-1}\bar S \phi(s_t, a_t) = Sf(s_t,a_t)
\end{align}
Using this representation, we can \emph{learn} the linear part of $f$,
$SN\bar S^{-1}$, directly from our state estimates: we just do a
linear regression from $\phi(Ss_t,a_t)$ to $Ss_{t+1}$.

For convenience, we summarize the entire learning algorithm (state
space discovery followed by system identification) as
Algorithm~\ref{alg:SSID-appendix}. 

\subsubsection{Filtering with the Extended Kalman Filter}\label{sec:EKF}
Whether we learn the dynamics through system identification or simply
write them down in the interpretable version of our state space, we
will end up with a transition model of the form~(\ref{eq:philintrans})
and an observation model of the form~(\ref{eq:linobsmod}).  Given
these models, it is easy to write down an EKF which tracks the robot
state.  The measurement update is just a standard Kalman filter update
(see, e.g.,~\cite{Thrun2005}), since the observation model is linear.
For the motion update, we need a Taylor approximation of the expected
state at time $t+1$ around the current MAP state $\hat s_t$, given the
current action $a_t$:
\begin{align}
s_{t+1}-s_t &\approx N [\phi(\hat s_t,a_t) +
\textstyle\frac{d\phi}{ds}\big|_{\hat s_t}(s_t-\hat s_t)]\\
{\textstyle\frac{d\phi}{ds}}\big|_{\hat s} &= 
(\hat s \otimes I + I \otimes \hat s) \otimes \bar a_t \otimes \bar a_t
\end{align}
We simply plug this Taylor approximation into the standard Kalman
filter motion update (e.g.,~\cite{Thrun2005}).

\clearpage
\begin{algorithm}[H] 
\caption{Robot System Identification}
\label{alg:SSID-appendix}
\textbf{In}: $T$ \emph{i.i.d.} pairs of observations $\{ {o}_t,
{a}_t\}_{t=1}^T$, measurement model for 4 landmarks ${C}_{1:4}$ (by e.g. GPS) \\ 
\textbf{Out}:  measurement model $\widehat C$, motion model $\widehat N$, robot states $\widehat X$ (the $t$th column is state $s_t$)\\
  \begin{algorithmic}[1]
\STATE Collect observations and odometry into a matrix $\widehat Y$ (Eq.~\ref{eq:Y}) 
\STATE  Find the the top
$7$ singular values and vectors: $\langle \widehat U, \widehat \Lambda, \widehat
V^\top \rangle \leftarrow \text{SVD}(\widehat Y,7)$
 \STATE Find the transformed measurement matrix $\widehat CS^{-1} = \widehat U $ and robot states $S\widehat X =  \widehat \Lambda \widehat V^\top$
 \STATE Compute a matrix $\Phi$ with columns $\Phi_t = \phi(Ss_t,a_t)$.
 \STATE Compute dynamics: $S\widehat N\bar S^{-1} = S\widehat X_{2:T} ( \Phi_{1:T-1})^\dag$
 \STATE Compute the partial $S^{-1}$: $\widehat S^{-1} =  {C}_{1:4}^{-1}(\widehat C_{1:4}S^{-1})$ where $\widehat CS^{-1}$ comes from step 3. $\widehat S^{-1} \widehat X$ gives us the $x,y$ coordinates of the states. These can be used to find $\widehat X$ (see Section~\ref{sec:orientation})
\STATE Given $\widehat X$, we can compute the full $S$ as $S = (S\widehat X) \widehat X^\dag$
 \STATE Finally, from steps 3,5, and 7, we find the interpretable measurement model $(\widehat C S^{-1})S$ and motion model $N = S^{-1}(SN\bar S^{-1})\bar S$. 
  \end{algorithmic}
  \end{algorithm}

\end{document}